\documentclass[twoside]{article}

\usepackage[accepted]{aistats2023}
\usepackage{epsfig}
\usepackage{graphicx}
\usepackage{amsmath}
\usepackage{amssymb}
\usepackage{booktabs}
\usepackage{wasysym}
\usepackage{subcaption}
\usepackage{multirow}
\usepackage{wrapfig}
\usepackage{enumitem}
\usepackage{xr}
\usepackage[mode=buildnew]{standalone}
\usepackage[round]{natbib}

\begin{document}

\twocolumn[

\aistatstitle{Deep Neural Networks with Efficient Guaranteed Invariances}

\aistatsauthor{Matthias Rath$^{\star,\dagger}$ \And Alexandru Paul Condurache$^{\star,\dagger}$}

\aistatsaddress{$^{\star}$Cross-Domain Computing Solutions \\ Robert Bosch GmbH \\ Stuttgart, Germany \And  $^{\dagger}$Institute for Signal Processing \\ University of L\"ubeck \\L\"ubeck, Germany} ]

\begin{abstract}
	We address the problem of improving the performance and in particular the sample complexity of deep neural networks by enforcing and guaranteeing invariances to symmetry transformations rather than learning them from data. Group-equivariant convolutions are a popular approach to obtain equivariant representations. The desired corresponding invariance is then imposed using pooling operations. For rotations, it has been shown that using invariant integration instead of pooling further improves the sample complexity. In this contribution, we first expand invariant integration beyond rotations to flips and scale transformations. We then address the problem of incorporating multiple desired invariances into a single network. For this purpose, we propose a multi-stream architecture, where each stream is invariant to a different transformation such that the network can simultaneously benefit from multiple invariances. We demonstrate our approach with successful experiments on Scaled-MNIST, SVHN, CIFAR-10 and STL-10. 
\end{abstract}

\section{\uppercase{Introduction}}
\begin{figure*}[t]
	\centering
	\includegraphics[width=0.9\linewidth]{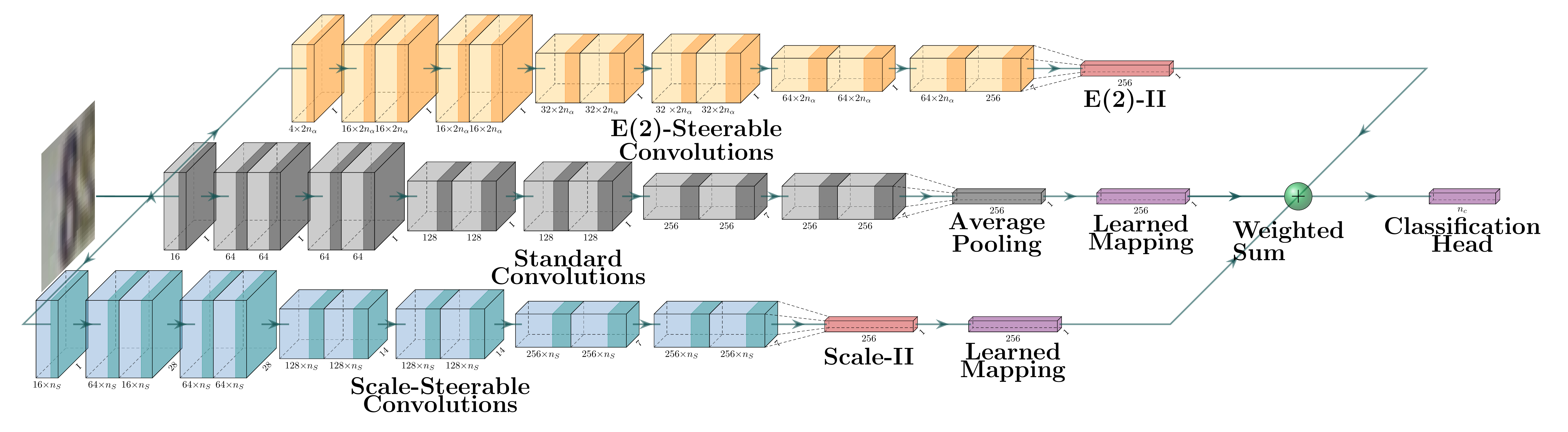}
	\caption{Triple-stream invariant Wide-ResNet16-4 architecture. Includes standard convolutions (grey), rotation-flip-steerable convolutions (E(2), orange), scale-steerable convolutions (blue), invariant integration layers (red), a weighted sum (green) and fully connected layers (purple). Residual shortcut connections are omitted for clarity.}
	\label{fig:Architecture}
\end{figure*}

Deep Neural Networks (DNNs) are one of the core drivers of technological progress in various fields such as speech recognition, machine translation, autonomous driving or computer vision \citep{DeepLearning}. At the core of their success lies the ability to process large amounts of data to yield solutions with remarkable generalization properties \citep{GeneralizationSurvey}. However, in many 
practical applications, the data is expensive to collect, store and label. Furthermore, it is rather difficult for humans to understand how such correlation-based methods work, which is a necessary first step in optimizing or adapting them to new application domains. Additionally, a certain degree of understanding and trust in the DNN's output is essential, e.g., when safety plays a major role \citep{Gran}.

Often, \textit{prior knowledge} about transformations that modify the desired output in a predictable way is available before training. Leveraging prior knowledge increases the interpretability of DNNs while also improving the sample complexity -- hence reducing the amount of data needed to obtain a desired performance. When incorporating this valuable prior knowledge into deep learning architectures, it is advantageous to \textit{guarantee} the corresponding in- or equivariances. We differentiate between \emph{invariance}, which is 
the property of a map to yield the same output for a transformed input and \emph{equivariance} which is 
the property of a map to preserve the transformation of the input, such that the output is transformed predictably.
One common example of leveraging equivariance in DNNs are convolutional layers \citep{Fukushima,LeCun}. These are equivariant to translations and can be easily used to generate a translation-invariant representation (e.g. by pooling).

For many tasks, we can define a set of \textit{symmetry transformations} that affect the desired output in a predictable way. 
For example, in the case of image classification, symmetry transformations of an input object map it within the same image class and thus do not change the desired classification output. Typical examples are rotations, translations and scales. Enforcing invariance to such transformations as an inductive bias decreases the sample complexity by reducing the search space while training a DNN. Moreover, guaranteed invariances contribute towards gaining an intuition on how inference is conducted in the DNN. \cite{GroupEquivariantCNNs} first used \textit{group equivariant convolutions} (G-Convs) in DNNs to enforce equivariance to transformations such as rotations and flips \citep{GroupEquivariantCNNs,HarmonicNetworks,SFCNN, E2STCNNs} or scales \citep{XuScale,LocalScaleInvariance,Sosnovik,DISCO}. For classification DNNs, those equivariant layers are usually followed by a max pooling operation among the group and spatial dimensions to obtain invariant features that are processed by the final classification layers. While max pooling guarantees invariance, it destroys important information that could be leveraged by a classifier and therefore lacks efficiency.
Since the transfer from equi- to invariant features has not been extensively investigated, it promises further improvement capabilities. 

Invariant Integration (II) is an algorithm to construct a complete feature space with respect to (w.r.t.) a transformation group \citep{SM_Existence}. So far, II has been used to replace the global spatial pooling operation for rotation-invariant classification DNNs. This resulted in an improved sample complexity by efficiently leveraging the available prior knowledge while adding targeted model capacity \citep{Rath,Rath2}. However, II has not been extended to other relevant symmetry transformations such as \textit{scales}. 

Group-equivariant DNNs usually incorporate prior knowledge about a single transformation. It is an open challenge how to proceed when \emph{multiple symmetries} are involved because it may be impossible to solve the constraints needed to design transformation-steerable filters depending on the involved groups. Even when avoiding the constraints via interpolation methods, simply expanding the regular equivariant G-Convs is computationally inefficient since the representation grows multiplicatively. For example, for a kernel with 8 rotations and 4 scales, we would have to store $8\cdot 4=32$ responses per kernel.

In this contribution, we extend the II framework beyond rotations and efficiently apply it to multiple transformations at once via a multi-stream architecture. Our \textbf{core contributions} are:
\begin{itemize}
	\item We adapt rotation-II to also include flips to achieve \textbf{invariance} to the 2D Euclidean group \textbf{E(2)}. 
	\item We \textbf{expand II towards scales}, thus covering a larger set of symmetry transformations. \item We address the issue of \textbf{multiple invariances} within a single architecture that effectively \textbf{combines} several streams, each one with \textbf{specific invariances} (see Figure \ref{fig:Architecture}). This significantly \textbf{extends the practical applicability} of \textbf{II}.
	\item We \textbf{evaluate} our approach on \textbf{Scaled-MNIST} and on the \textbf{real-world datasets} SVHN, CIFAR-10 and STL-10. On STL-10 using only labeled data, we report new \textbf{state-of-the-art} results. 
\end{itemize}

\section{\uppercase{Related Work}}
\subsection{Group-Equivariant Neural Networks}
\textit{Group-equivariant convolutional} layers were proposed by \cite{GroupEquivariantCNNs} and applied to discrete 90$^\circ$ rotations and flips by transforming the filters and storing all responses among a \textit{group channel}. This approach uses the \textit{regular} group-representation. Extensions apply this principle to finer-grained rotations via interpolation \citep{Bekkers,Hoogeboom}, rotation-steerable filters \citep{SFCNN,E2STCNNs} or by learning all rotated versions of a filter with invariant coefficients \citep{Diaconu1}. The maximum response can be stored as the orientation in a vector field \citep{MarcosRot}. This is closely related to the \textit{irreducible representation} which achieves continuous rotation-equivariance via complex-valued responses \citep{HarmonicNetworks}. 
In general, it has been proven that G-Convs are the most general equivariant linear map and a necessary condition for equivariant DNNs \citep{Kondor,GeneralEquivariantCNNs,EstevesSurvey}. 

Besides rotations, in- or equivariance to scale transformations plays a major role in many practical applications. In- or equivariance can again be achieved by sharing filters among different scales using bi-linear interpolation \citep{XuScale}, scaling the input \citep{LocalScaleInvariance} or processing the maximum response and the corresponding scale as a vector field \citep{MarcosScale}. Filters for scale-equivariant convolutions can be constructed using scale-steerable filters with log-radial harmonics \citep{GhoshScale}, Hermite polynomials \citep{Sosnovik}, optimized discrete bases \citep{DISCO}, or with separable Fourier-Bessel bases \citep{ZhuScale}. A scale-equivariant G-Conv operating on scale-spaces was introduced in \citep{Worrall19}.

Other work investigates general input domains such as the 3D Euclidean space \citep{CubeNet,3DSTCNNs,cesa2022}, spheres \citep{SphericalCNNs,SphereNet,SpinWeightedSpherical,DefferrardDeepSphere,ClebschGordanNets,Jiang,Shakerinava}, general manifolds \citep{Gauge,Finzi2020}, or general groups \citep{BekkersLie,Finzi2}. Moreover, equivariant versions of non-linear maps such as attention and transformers have been introduced \citep{Fuchs,IterativeSE3Transformer,Romero,Romero1,GroupSelfAtt,LieTransformer,GaugeEquivariantTransformer}. The non-linearities and sub-sampling layers used in equivariant CNNs have been investigated in \citep{EquivNonlin,GroupSubsampling}.

\subsection{Invariant Neural Networks}
When solving tasks that require invariance, group-equivariant DNNs are usually followed by a global (max) pooling layer among the group and spatial dimensions. While max pooling guarantees invariance, it is affected by a loss of information, since all but the maximum value are discarded. Obtaining invariance in a more sophisticated way thus promises to further improve invariant DNNs.

One approach to achieve invariance while also ensuring separability is \textit{Invariant Integration} (II), proposed by \cite{SM_Existence,SM_Algos}. II is an algorithm to construct a complete feature space w.r.t. a group, i.e., similar patterns are mapped to the same point while distinct patterns are mapped to distinct points. II has been used in combination with conventional machine learning classifiers for image classification \citep{SM_Gray}, event detection within a cascaded feature extractor to obtain invariance to anthropometric changes \citep{Condurache} or robust speech recognition \citep{Muller1,Muller2,Muller3}. Rotation-II has been used to replace the global spatial pooling layer within rotation-invariant deep learning architectures \citep{Rath,Rath2} and shown to further increase the sample complexity of such networks. \cite{Puny} solve the group average, which II is based on, for larger, intractable groups by integrating over a subset. They applied their method to classification for motion-invariant point clouds and graph DNNs integrating over the whole DNN. The invariant integral has also been used to prove that in- and equivariance improve generalization when the target distribution is in- or equivariant \citep{Elesedy}.

Whereas \cite{Rath,Rath2} focused on II for the group of rotations, we expand this framework to scales as well as flips (using E(2)). Thereby, we show that the framework can be expanded to general group transformations and generally improves the sample complexity of group-equivariant CNNs in classification tasks. Most related work focuses on single transformation groups. Through our multi-stream architecture, we propose a novel approach that allows our network to learn the best possible combination of invariant features among multiple transformations at once. Another method that combines equivariance to both rotations and scales is the Polar Transformer Network (PTN) which process inputs in the polar coordinate system \citep{PolarTN}. However, working in polar coordinates, although advantageous for rotations, may prove to be detrimental to translation equivariance. Indeed PTNs are by design invariant to translation, but it is not clear how much other relevant information is destroyed.
At the same time, Spatial Transformer Networks \citep{STN} in general cannot offer invariance guarantees, as they rely on the localization network to learn the correct transformation. The invariance is not fully guaranteed but approximated w.r.t the estimated transformation (STNs) or object center (PTNs). The same is valid for deformable \citep{DeformableConvs} and tiled convolutions \citep{TiledConvs}. 

\section{\uppercase{Theoretical Background}}
\subsection{In- and Equivariance} 
In- and equivariance are mathematical concepts describing the behavior of a map $f: \mathbb{R}^n \to \mathbb{R}^m$ under transformations of the input that can be modeled using the mathematical abstraction of a group. A group is a set $G$ equipped with a group operation $\cdot : G\times G \to G$ fulfilling the four group axioms: closure, associativity, identity and invertibility. The map $f$ is called equivariant w.r.t. $G$, if left group actions $L_gx$ acting on the input $x \in \mathbb{R}^n$ result in predictable changes $L_{g^\prime}$ of the output
\begin{equation}
\small{\forall x \; \forall g \; \exists g^\prime \; \text{s.t.} \; f(L_gx) = L_{g^\prime} f(x) \text{,}}
\end{equation}
where the left group action $L_g : \mathbb{R}^n \rightarrow \mathbb{R}^n$ is defined for each group element $g \in G$. The group elements $g$ and $g^\prime$ are not necessarily equal, i.e., the transformation of the output can be different from the one applied on the input, but is predictable. 
If the output does not change, 
i.e. $\forall x \; \forall g \text{, } f(L_gx) = f(x)$, the function is called invariant.

In the context of CNNs, the network and the layers are maps between feature spaces that can be described by $f: \mathbb{Z}^2\to \mathbb{R}^n$. The effect of input transformations on the feature maps and outputs can thus be studied using Group Theory. In the course of the paper, we use left actions on the input $x$ by lifting the left group action of $G$ on $\mathbb{Z}^2$ via $L_gx(y) = x(g^{-1}y)$ where $y \in \mathbb{Z}^2$ are the pixel coordinates. The transformed input is equivalent to the original input at the point $g^{-1}y$ that gets mapped to $y$ by $g$.

\subsection{Group-Equivariant Convolutions}
Group-equivariant convolutions are the most general linear map achieving equivariance to a transformation group $G$ \citep{Kondor,GeneralEquivariantCNNs,EstevesSurvey}. \cite{GroupEquivariantCNNs} first used G-Convs in the context of DNNs to learn representations with guaranteed equivariance to group transformations. The continuous G-Conv of two functions $f$ and $\psi$ is defined as
\begin{equation}\label{eq:groupConv}
\small{(f \star_G \psi)(u) = \int_{g\in G}f(g)\psi(u^{-1}g)d\mu(g) \text{,}}
\end{equation}
where $d\mu(g)$ is the Haar measure with $\int_{g\in G}d\mu(g) = 1$. The in- and output are defined on the group itself with $g, u \in G$. The \textit{standard convolution} is a special case of G-Convs where the elements are given by $y, t \in \mathbb{Z}^2$ and the inverse action $t^{-1}y$ results in shifts $y-t$. A DNN is group equivariant, if and only if each of its mappings is equivariant to or commutes with the group \citep{Kondor}. In many cases, the transformation group $G$ can be split into the translation group T defined on $\mathbb{Z}^2$ and the corresponding quotient group $H = G/T$. In order to achieve equivariance to $G$, a $H$-equivariant convolution can be applied at all spatial locations of the input using a standard convolution. Different representations can be used to compute and store the equivariant features. \textit{Irreducible} representations require the minimal possible number of parameters, but often involve complex calculations. For 2D rotations, using irreducible representations results in complex-valued feature spaces where the orientation information is stored via the complex phase. The \textit{regular} representation of a discrete group stores all responses of transformed filters among an additional channel called \textit{group channel}.

To compute regular G-Convs, either the input or the filters need to be transformed for all $g \in G$. A simple method is to transform the filters using interpolation methods such as bi-linear interpolation. However, this introduces sampling artifacts and consequently weakens the equivariance guarantees.
An alternative approach is to use transformation-steerable filters, which have been first introduced for rotations \citep{Adelson}. While steerable filters introduce a computational overhead compared to simpler interpolation methods, arbitrarily transformed versions can be calculated in closed-form and are thus not afflicted by sampling effects. This concept can be effectively used for G-Convs by restricting the learned filters to linear combinations of steerable basis filters.
\cite{SFCNN} built steerable filters for rotation G-Convs using a Gaussian kernel which \cite{E2STCNNs} expanded to the general E(2)-group. \cite{Sosnovik,DISCO} constructed scale-steerable filter CNNs using 2D Hermite polynomials or a learned discrete basis. We use the state-of-the-art methods for rotation- (E(2)-STCNNs, \citealt{E2STCNNs}) and scale-invariant (DISCO, \citealt{DISCO}) tasks as our baseline.

\subsection{Invariant Representations in DNNs}
Invariance plays a major role in many DNN applications. 
For the example of classification, input transformations that do not change the desired class output should not change the learned feature space. Group-equivariant DNNs typically use pooling to transfer from equi- to invariant representations. When operating on regular G-Convs, the pooling procedure is two-fold: Pooling among the transformation channel creates an equivariant representation where an input-transformation induces the same transformation in the feature space; and pooling among the spatial dimension obtains the final invariance.
For closed groups, such as rotations and flips, spatial average or max pooling can be used to obtain invariant representations. This is different for the scale group, where average pooling over the spatial dimension does not lead to invariant, but rather homogeneous features, i.e., scaling by $s$ modifies the output by multiplying with $s^2$.

\subsection{Invariant Integration}
Invariant Integration is an algorithm to construct a complete feature space $\mathcal{F}$ w.r.t. a transformation group $G$ introduced in \cite{SM_Existence}. A feature space is complete, if all equivalent patterns w.r.t. the transformation are mapped to the same point while all distinct patterns are mapped to different points. For this mapping, II uses the group average $A[f](x)$ which integrates over all possible transformations $g\in G$ of an input $x$ processed by a polynomial $f$
\begin{equation}
\small
{A[f](x) = \int_{g\in G}f(L_gx)d\mu(g) \text{.}}
\end{equation} 
In our case, $x$ is the output of the final feature map after pooling among the group dimension. For $f$, \cite{SM_Algos,SM_Gray} used the set of monomials $m(x) = \prod_{i=1}^{M}x_i^{b_i}$ with $\sum_i b_i \leq |G|$, defined as a product of individual signal values $x_i$ with exponents $b_i$, which have been shown to be a good choice to maintain a high expressiveness of the invariant features \citep{Noether1916}. For 2D rotations and translations, II with monomials within a local neighborhood defined by distances $d_i$ results in
\begin{equation}\label{eq:II_Mon}
\small{
	A[\text{m}](x) = \frac{1}{N_\phi UV}\sum_{\phi, t}\prod_{i=1}^{M} x[t-L_\phi d_i]^{b_i}} \text{.}
\end{equation}
\cite{Rath} applied II on top of equivariant G-Convs within a DNN. The II layer is differentiable, if $x_i > 0$ $\forall i$, which allows for an end-to-end optimization of the DNN via backpropagation. To select a meaningful subset of monomials, an iterative algorithm based on the least square solution of a linear classifier can be used. However, a pruning-based approach leads to a more streamlined training procedure \citep{Rath2}. 
Additionally, selecting the monomials can be avoided by replacing them with alternatives well-known in deep learning literature such as self-attention, a weighted sum (WS) or a multi-layer-perceptron. For rotation-II, the special case using a WS achieves the best performance while being easier to train \citep{Rath2}. II using a WS with a learnable kernel $\psi \in \mathbb{R}^{k\times k}$ applied to $N_\phi$ finite rotations $\phi \in [0^\circ, \frac{360^\circ}{N_\phi}, \hdots]$ obtained using bi-linear interpolation with input dimensions $U \times V$ is defined as
\begin{equation}\label{eq:II_Weighted}
\small{A[\text{WS}](x)
	= \frac{1}{N_\phi UV}\sum_{\phi, t}\sum_{y \in \mathbb{Z}^2} x(y)L_\phi\psi(y-t)} \text{.}
\end{equation}

\section{\uppercase{Method}}
In this section, we extend rotation-II using a WS to the E(2)-group involving rotations and flips (Section \ref{sec:E2II}). We then show why II cannot be straightforwardly applied to scale transformations and introduce an alternative to obtain scale-invariants based on II (Section \ref{sec:ScaleII}). We then propose a scale-invariant CNN including Scale-II (Section \ref{sec:ScaleApp}). Finally, we introduce a multi-stream DNN architecture that efficiently combines invariances to multiple transformations within a single DNN (Section \ref{sec:MultiStream}).

\subsection{E(2)-Invariant Integration}\label{sec:E2II}
Equation \ref{eq:II_Weighted} can straightforwardly be expanded to each discrete subgroup of E(2) (as used by \citealt{E2STCNNs}), which contains flips $L_f\psi(y)$ in addition to rotations
\begin{equation}\label{eq:II_WeightedE2}
\small{A[\text{WS}](x) 
	= \frac{1}{2N_\phi UV}\sum_{f,\phi,t}\sum_{y \in \mathbb{Z}^2} x(y)L_\phi L_f\psi(y-t)} \text{.}
\end{equation}

\subsection{Scale-Invariant Integration}\label{sec:ScaleII}
In images, objects naturally appear at different scales, e.g., due to variable camera-to-object distances. Hence, DNNs for object classification or detection benefit from invariance to scales. We propose to use II in combination with a scale-equivariant CNN to obtain scale-invariant features.
In comparison to the rotation group, discrete scale transformations are not circular and non-invertible due to the loss of information (e.g. during down-scaling). Thus, scales do not satisfy all group axioms and can only be modeled as a semi-group.
\cite{SM_Existence} demonstrated that it is impossible to construct invariants by integrating over the scale semi-group while at the same time achieving separability. This prohibits constructing a complete feature space w.r.t scales using the standard II approach. Nevertheless, the group average w.r.t. translations is a homogeneous function w.r.t. scales when using polynomials \citep{SM_Algos}. This means that the effect of scaling by $s$ on the features obtained using translation-II is defined as
$A[f](L_sx) = s^K A[f](x)$
where the order $K$ is defined by the polynomial order of $f$ with the scale operator $L_s[f](y) = f(s^{-1}y) \; \forall s > 0$.

As shown by \cite{SM_Algos}, a complete feature space w.r.t. the scale-translation semi-group $G_S = S\rtimes T$ can be calculated by dividing homogeneous functions of the same order. When using monomials $m$, one resulting scale-invariant integral is given by dividing monomials of the same order $\sum b_{1,i} = \sum b_{2,i}$ with $t\in \mathbb{Z}^2$
\begin{equation}\label{eq:ScaleIIMon}
\small
{A_{G_S}[m](x) =\frac{A_T[m_1](x)}{A_T[m_2](x)} = \frac{\sum_{t}\prod_{i}x(t-d_{1,i})^{b_{1,i}}}{\sum_{t}\prod_{i}x(t-d_{2,i})^{b_{2,i}}}} \text{.}
\end{equation}
The special case of translation-II that combines values within a fixed neighborhood using a learnable WS as function $f$ results in a standard convolution followed by Average Pooling and is equivalent to a polynomial of order 1. Consequently, we can choose a divisor of the same homogeneous order to obtain a scale-invariant representation and use the mean of the feature map.
We thus introduce the WS-based scale-group average $A_{G_S}$ based on Average Pooling over a standard convolution without bias with translations $L_t$, $y,t \in \mathbb{Z}^2$ divided by the mean
\begin{equation}\label{eq:ScaleII}
\small{
	A_{G_S}[\text{WS}](x) = \frac{A_T[\text{WS}](x)}{\sum_{y}x(y)}
	= \frac{\sum_{y}\sum_{t}x(y)\psi(y-t)}{\sum_{y}x(y)}} \text{.}
\end{equation}

\noindent\textbf{Proof of Invariance.}
Since translation-II and the mean are both homogeneous w.r.t. scales with factor $s^2$, it is easy to see that dividing them leads to a scale-invariant solution
\begin{equation}
\small{
	\frac{A_T[f](L_sx(y))}{\sum_{y}L_sx(y)} = \frac{s^2 A_T[f](x)}{s^2 \sum_yx(y)} = \frac{A_T[f](x)}{\sum_{y}x(y)}} \text{.}
\end{equation}

\subsection{Application to DNNs}\label{sec:ScaleApp}
Inspired by the work on rotations in \cite{Rath, Rath2}, we apply E(2)- and Scale-II (Formulas \ref{eq:II_WeightedE2} - \ref{eq:ScaleII}) on top of the corresponding equivariant features learned using regular G-Convs with steerable filters: E(2)-STCNNs \citep{E2STCNNs} and DISCO \citep{DISCO}. Those features are processed via max pooling among the group dimension. II then replaces the spatial pooling operation to obtain invariant features that are in turn processed by the final classification layers.
For E(2)-II, we use the WS approach. For Scale-II, we investigate both proposed variants. For Scale-II with monomials, we follow \cite{Rath2} and randomly select monomials that are iteratively pruned during training to find the most relevant ones. We ensure the same monomial order between dividend and divisor by normalizing the divisor's exponents with $\frac{\sum_{i}b_{1,i}}{\sum_{i}b_{2,i}}$. Additionally, $x_i>0$ ensures a differentiable solution. For Scale-II with WS, each convolution depends on all input channels. Consequently, we divide by the mean over all input channels for the WS-based II. Moreover, it is important that $\sum_y x(y) > 0$. Hence, we apply both II variants to the output of a ReLU layer minimized to a small value $\epsilon>0$. 

\subsection{Multi-Stream Invariance}\label{sec:MultiStream}
To obtain features with guaranteed invariance to multiple transformations, we implement a DNN architecture with multiple streams. Each stream is invariant to a dedicated transformation enforced using G-Convs and II (see Figure \ref{fig:Architecture}). While multiple transformations could in theory be embedded into a single network, the regular representations the network needs to process would grow with $\mathcal{O}(N\cdot M)$ for group sizes $N$ and $M$. In contrast, using a dedicated stream per transformation only increases the number of representations by $\mathcal{O}(N+M) \ll \mathcal{O}(N\cdot M)$. 

The input is processed separately by each transformation-invariant stream using G-Convs with steerable filters. II is applied on top of the learned scale- and E(2)-equivariant features to obtain invariant representations. For the standard convolution stream (std.) we use average pooling. Thus, we have $\mathbf{x}_{j} \in \mathbb{R}^{C_j}$ with $j \in \{\text{e2}, \text{scale}, \text{std}\}$. We combine two or three streams, each one invariant to either rotations and flips, scales or translations (std.), using two steps. First, we map all features to the same dimension using a linear map $\mathbf{\tilde{x}}_{j} = \mathbf{W_{j}\mathbf{x}_j}$ with $\mathbf{W_{j}} \in \mathbb{R}^{C_\text{map} \times C_j}$. We then combine these streams via a normalized learnable WS, e.g. for the case of three streams $\mathbf{x}_{\text{combined}} = \mathbf{w}_{\text{e2}} \circ\mathbf{\tilde{x}}_{\text{e2}} + \mathbf{w}_{\text{scale}}\circ \mathbf{\tilde{x}}_{\text{scale}} + \mathbf{w}_{\text{std}}\circ \mathbf{\tilde{x}}_{\text{std}}$ with $\mathbf{w}_j \in \mathbb{R}^{C_\text{map}}$ initialized to 1 and normalized s.t. $(\mathbf{w}_{\text{e2}}+\mathbf{w}_{\text{scale}}+\mathbf{w}_{\text{std}})_i=1$ with $i = 1, \hdots, C_\text{map}$ and the Hadamard product $\circ$ inspired by the learnable channel-wise scaling used in BatchNorm layers \citep{BatchNorm}. This approach allows to combine the invariant features with an explicitly learned factor s.t. the network can learn, which invariances are the most relevant for the corresponding task. 

Each stream is pre-trained individually. We then combine the architecture, freeze all convolutional weights and only train the linear maps $\mathbf{W}_\text{j}$, the WS $\mathbf{w}_{j}$ and the classification weights $\mathbf{w}_{\text{out}} \in \mathbb{R}^{C_{\text{map}} \times n_c}$ with $n_c$ classes. The combination head worked best, when mapping the other streams to the E(2)-output, i.e., $C_\text{map}=C_\text{e2}$ and keeping $\mathbf{W}_\text{e2}$ fixed as the identity. Appendix \ref{sec:Combination} provides results with different combinations, e.g., mapping all streams or concatenation. Our dedicated training procedure allows to fine-tune each stream individually, which provides a good initialization point for the combination head and further improves the sample complexity compared to a full end-to-end training (cf. Appendix \ref{sec:AblFull}). While this causes an overhead at train time, all operations can easily be fused into a single network at inference time. 

\section{\uppercase{Experiments \& Discussion}}
The proposed scale-II algorithm is evaluated on Scaled-MNIST \citep{MNIST-Scale} and the invariant multi-stream networks on SVHN \citep{SVHN}, CIFAR-10 \citep{CIFAR-10} and STL-10 \citep{STL10}. We use the full dataset and limited subsets with $N_t$ samples to assess the sample complexity. The subsets are sampled with constant class balance and are the same for all variants. For Scaled-MNIST, we use $N_{t} \in \{10,50,100,500,1k,5k,10k,12k\}$. For SVHN and CIFAR-10 we use $N_{t} \in \{100,500,1k,5k,10k,50k\}$.

For SVHN, CIFAR-10 and STL-10 we use Wide-ResNets (WRNs, \citealt{WideResNet}) as backbone. We use the respective \textit{standard data augmentations}: scales for Scaled-MNIST, no augmentations for SVHN, shifts and crops for CIFAR-10, and shifts, crops and Cutout \citep{Cutout} for STL-10. For all single-streams, the number of trainable parameters is constant. We report results for the multi-stream architectures with full streams and constant number of parameters. We optimize all hyper-parameters (HPs) using a 80:20 validation split and Bayesian Optimization with Hyperband (BOHB, \citealt{BOHB}). For all II-WS-layers, we use $k=3$ and a constant number of in- and output channels. For II with monomials, we use an iterative pruning-based selection with $n_m=\{25,12,5\}$ monomial pairs after 0, 5 and 10 epochs following \cite{Rath2}. The exact HPs, optimization settings and network architectures can be found in Appendix \ref{sec:HPs}. If not mentioned otherwise, we report the mean and standard deviation over three runs.

\subsection{Evaluating Scale-Invariant Integration}
\begin{figure}[t]
	\centering
	\includegraphics[width=0.9\columnwidth]{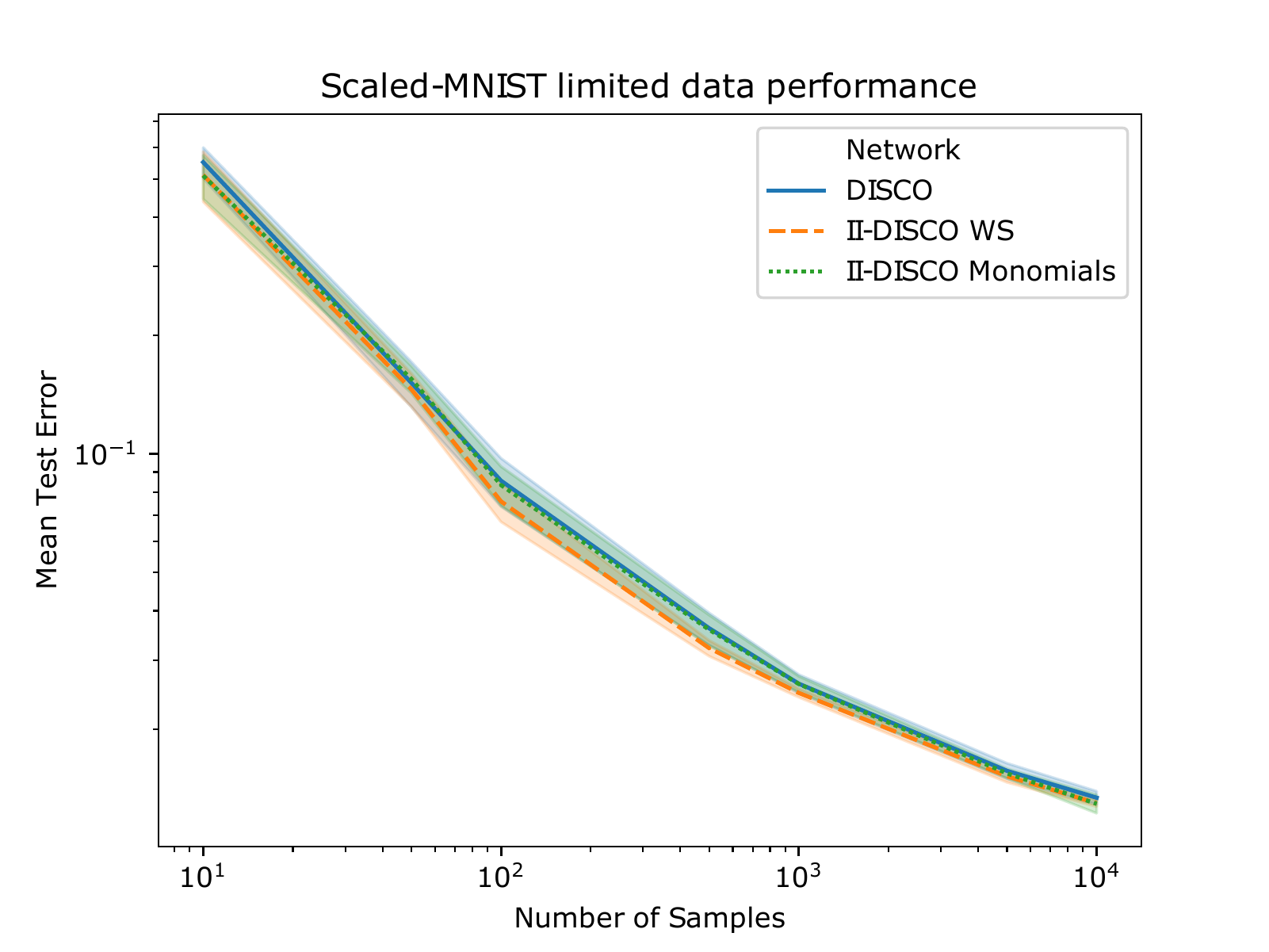}
	\caption{Log. Test Error (TE) on Scaled-MNIST subsets.}
	\label{fig:limited_MNIST}
\end{figure}

\begin{figure*}
	\hfill
	\begin{minipage}[t]{0.95\columnwidth}
		\small
		\centering
		\captionof{table}{Invariance Error $\Delta$ when obtaining a scale-invariant representation using the respective layer directly on the input and a randomly initialized scale-equivariant network.}\label{tab:InvarianceError}
		\begin{tabular}{ccc}
			\toprule
			Layer & Input & CNN \\
			\midrule
			Average Pooling & 0.222 & 0.090 \\
			Mixed Pooling & 0.017 & 0.039 \\
			\textbf{Scale-II} Monomials & 1.24e-4 & \textbf{2.64e-7} \\
			\textbf{Scale-II} WS & \textbf{2.97e-9} & 5.94e-3 \\ 
			\bottomrule
		\end{tabular}
	\end{minipage}
	\hfill
	\begin{minipage}[t]{0.95\columnwidth}	
		\centering
		\small
		\centering
		\captionof{table}{Test Error (TE) on Scaled-MNIST using a CNN with 5 layers, data augmentation with random scales and an upsampling layer to double the input size.}\label{tab:Scaled-MNIST}
		\begin{tabular}{ccc}
			\toprule
			Method & Scale-II & TE [\%] \\
			\midrule
			CNN & - & 1.60 $\pm$ 0.09\\
			SES-CNN & - & 1.42 $\pm$ 0.07\\
			DISCO & - &1.35 $\pm$ 0.05 \\
			\textbf{II}-DISCO & Monomials &\textbf{1.30} $\pm$ 0.06 \\
			\textbf{II}-DISCO & WS & \textbf{1.30} $\pm$ 0.02 \\
			\bottomrule
		\end{tabular}
	\end{minipage}
\end{figure*}

We evaluate our scale-II layer on the Scaled-MNIST dataset, which consists of hand-written digits artificially scaled with factor $s \in [0.3,1]$. We use the architecture from \cite{Sosnovik} (SES-CNN) and \cite{DISCO} (DISCO) built of three convolutional and two dense layers using $n_S=4$ scales and replace the final global pooling layer with the scale-II layer. Following \cite{Sosnovik,DISCO}, we compare the invariance error of the scale-II layers to the methods they use to obtain scale-invariant representations: a mixed pooling approach including average and max pooling for Scaled-MNIST and average pooling for STL-10. The invariance error is a simplified version of the equivariance error given as
$\Delta=\frac{1}{S} \sum_s \frac{|\psi(x)-\psi(L_s x)|_2^2}{|\psi(x)|_2^2}$.
We compute $\Delta$ when directly processing the input, i.e., $\psi$ is the II- or pooling-layer and when $\psi$ is the randomly initialized CNN including the respective layer using 100 samples from Scaled-MNIST scaled with $s\in [0.5, 0.55, ..., 1.0]$ .
The results in Table \ref{tab:InvarianceError} show that scale-II guarantees invariance as opposed to the pooling approaches. While Scale-II with WS achieves a better error directly on the input, the monomial variant is slightly better when applied within the DNN. 

We then evaluate the performance of our scale-II layer on Scaled-MNIST for classification. Here, scale-invariance is paramount to obtain correct results because the test set contains more variability than the training set, thus benefiting scale-invariant algorithms. 
Table \ref{tab:Scaled-MNIST} shows the results on the full dataset, Figure \ref{fig:limited_MNIST} with limited training data. We report mean and standard deviation over the six pre-defined data splits.
II outperforms the pooling approach on full data and in the limited sample regime highlighting the improved sample complexity of our approach. In summary, the mixed pooling approach used by \cite{DISCO} does not guarantee scale-invariance which leads to a decreased performance.  
II allows for invariance guarantees and the WS-variant is easier to optimize than the monomials-variant. Hence, II with a WS outperforms the latter in the limited data domain and is used for all further experiments.

\subsection{Multi-Stream Digit Classification}
\begin{figure*}
	\begin{minipage}[t]{.65\textwidth}
		\small
		\centering
		\captionof{table}{TE on SVHN, CIFAR-10 and STL-10. WRN16-4, WRN28-10 and WRN16-8 are used as baseline architecture. $^\star$ indicates constant number of parameters.
		}\label{tab:Results}
		\begin{tabular}{cccccc}
			\toprule
			II & Streams & Invariance & SVHN [\%] & C-10 [\%] & STL-10 [\%]\\
			\midrule
			\multicolumn{2}{c}{PTN } & Rot. \& Scale & 2.97 & 6.72 & - \\ 
			x & Single & Std.
			& 2.93 $\pm$ 0.02 & 3.89 $\pm$ 0.08 & 12.02 $\pm$ 0.05 \\ 
			x & Single & E(2) 
			& 2.64 $\pm$ 0.05  & 2.91 $\pm$ 0.13 & 9.80 $\pm$ 0.40 \\ 
			x & Single & Scale 
			& 2.71 $\pm$ 0.01 & 4.04 $\pm$ 0.03 
			& 8.07 $\pm$ 0.08 \\
			\midrule 
			\checkmark & Single & E(2) & 2.36 $\pm$ 0.05 & 2.95 $\pm$ 0.04 
			& 7.67 $\pm$ 0.07 \\ 
			\checkmark & Single & Scale & 2.54 $\pm$ 0.04 & 3.91 $\pm$ 0.12 
			& 7.92 $\pm$ 0.09 \\
			\checkmark & Dual$^\star$ & Scale \& E(2) & 2.20 $\pm$ 0.05
			& 2.96 $\pm$ 0.10 
			& 6.38 $\pm$ 0.15\\ 
			\checkmark & Dual & Scale \& E(2) & 2.12 $\pm$ 0.11 
			& 2.75 $\pm$ 0.04 
			& 5.95 $\pm$ 0.11 \\
			\checkmark & Triple$^\star$ & Scale, E(2) \& Std. & 2.29 $\pm$ 0.03 
			& 2.74 $\pm$ 0.08   
			& 6.46 $\pm$ 0.08 \\
			\checkmark & Triple & Scale, E(2) \& Std. & \textbf{2.10} $\pm$ 0.07 
			& \textbf{2.68} $\pm$ 0.03 
			& \textbf{5.90} $\pm$ 0.05 \\
			\bottomrule
		\end{tabular}
	\end{minipage}
	\hfill
	\begin{minipage}[t]{.3\textwidth}
		\small
		\centering
		\captionof{table}{Role of the II layer for our triple-stream network on STL-10. An 'x' marks an invariant stream without II. }\label{tab:AblMS}
		\begin{tabular}{ccc}
			\toprule
			\multicolumn{2}{c}{II} & \\
			\cmidrule{1-2}
			E(2) & Scale & TE [\%] \\
			\midrule
			x & x & 7.51 $\pm$ 0.12 \\ 
			x & \checkmark & 7.35 $\pm$ 0.09 \\ 
			\checkmark & x & 6.34 $\pm$ 0.07\\ 
			\checkmark & \checkmark & \textbf{5.90} $\pm$ 0.05 \\
			\bottomrule
		\end{tabular}
	\end{minipage}
\end{figure*}

\begin{figure*}[t]
	\centering
	\begin{minipage}{.48\textwidth}
		\centering
		\includegraphics[width=\columnwidth]{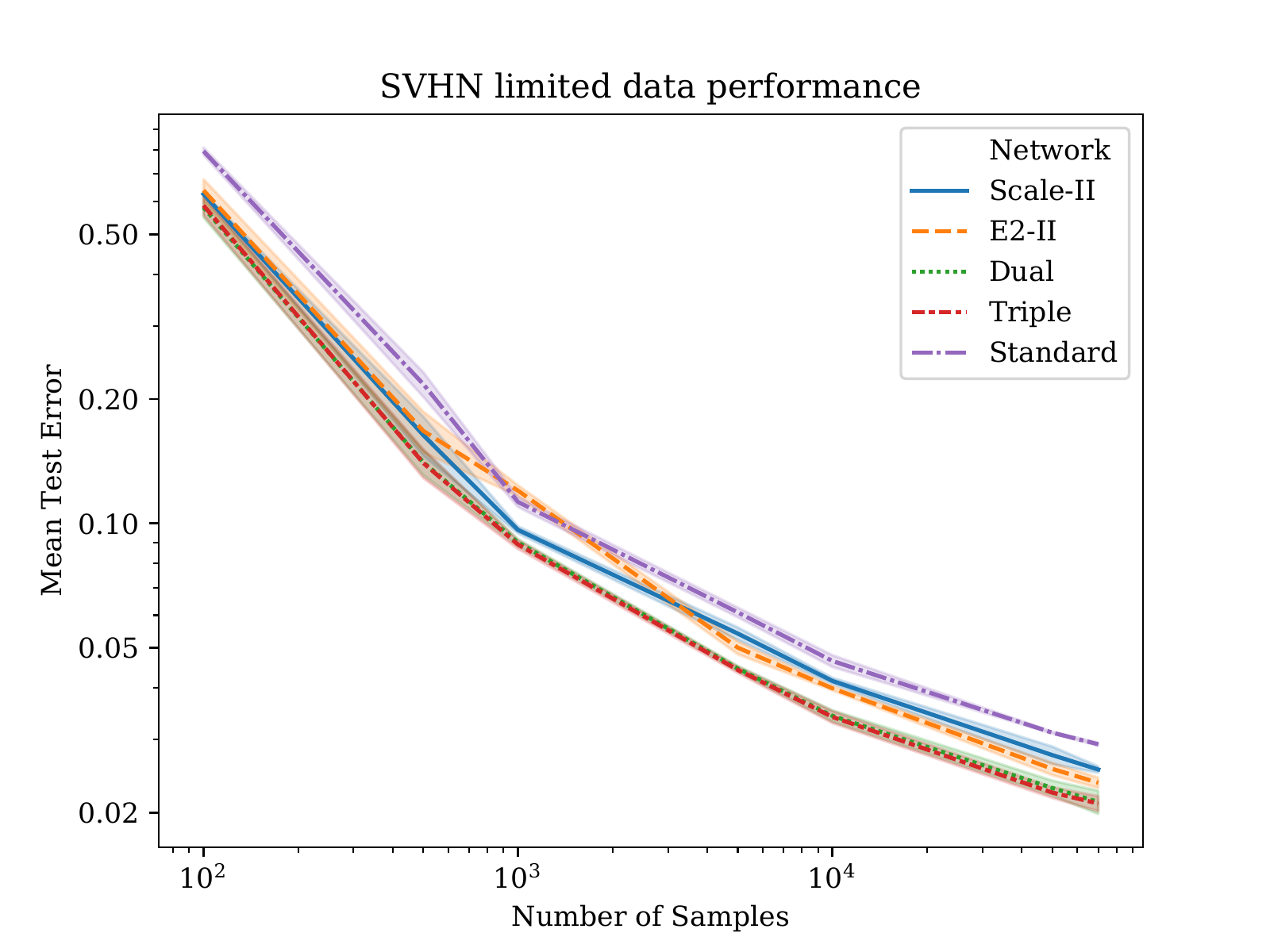}
		\captionof{figure}{Log. TE on subsets of SVHN with full streams.}
		\label{fig:limited_SVHN}
	\end{minipage}%
	\hfill
	\begin{minipage}{.48\textwidth}
		\centering
		\includegraphics[width=\columnwidth]{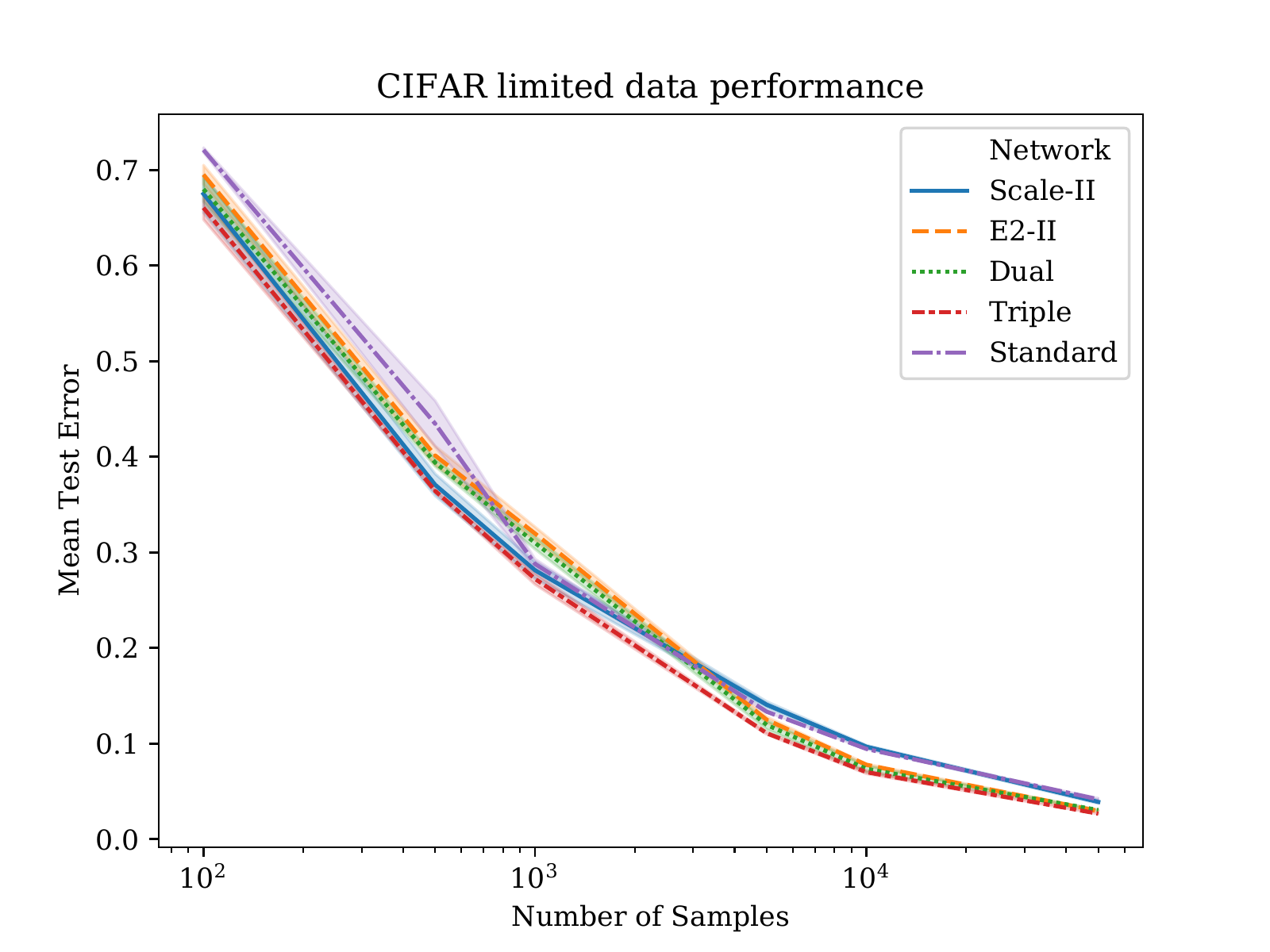}
		\captionof{figure}{TE on subsets of CIFAR-10 with full streams.}
		\label{fig:limited_CIFAR}
	\end{minipage}
\end{figure*}

The Street View House Number (SVHN) dataset contains single digits taken from house numbers. SVHN includes digits with different colors, font types, orientations and backgrounds and is thus harder to solve than MNIST. For all experiments on SVHN, we use the core training data, a WRN16-4 architecture, $n_r = 8$ rotations for all E(2)-G-Convs and $n_S = 3$ scales for the scale stream. The II step is calculated using the same number of rotations and flips.
The results on the full SVHN dataset are shown in Table \ref{tab:Results} while Figure \ref{fig:limited_SVHN} depicts the results using limited training data. 

Using II instead of pooling improves the accuracy for all invariant architectures showing that II better preserves the information when transferring to invariance. Both invariant single-stream networks outperform the std. CNN in the limited and full data domain. This indicates that invariance to rotations and flips as well as scales is valuable information the classifier needs to learn during training -- and the training data does not cover enough variability w.r.t rotations and scales for the baseline to learn this information.
The rotation- and flip-invariant DNN achieves a better sample complexity than the scale-invariant one, indicating that rotation-invariance is more valuable for this dataset. 

The multi-stream networks significantly outperform all single-stream variants including the baseline in all data regimes. The multi-stream architecture learns meaningful combinations of the generated invariants and is able to automatically choose the invariance best-fit for the training data at hand. This allows for best performances on all dataset sizes. 
In addition, for the rather simple task of classifying numbers, combining only a rotation- and a scale-invariant stream outperforms a standard CNN and leads to almost the same performance as with additional std. convolutions. In this problem setup, rotation- and scale-invariances seem sufficient for optimal performance and are able to recover all necessary object invariances. We conjecture that the learned features in the network's layers within the invariant streams focus on global invariances like illumination and noise while the dedicated II layer handles specific object invariances, e.g. to scales, similar to the scattering transformation \citep{Scattering}.

\subsection{Multi-Stream Object Classification}
Finally, we evaluate our proposed architecture for the more complex task of object classification on CIFAR-10 and STL-10. Both datasets contain RGB images of ten different object classes. STL-10 is a subset of ImageNet containing 5k training images that is commonly used as a benchmark for the limited data performance of object classification networks. STL-10 is more challenging than CIFAR-10 since it contains bigger and more diverse images. For CIFAR-10, we use WRN28-10 and the E(2)-stream with 8, 8 and 4 rotations per residual block. For STL-10, we use WRN16-8 and the E(2)-architecture with 8, 4 and 1 rotation per residual block \citep{E2STCNNs}. The E(2)-II-layer is used for 4-rotations and flips or flips-only, respectively. For both datasets, the scale-stream uses 3 scales \citep{DISCO}. The STL-10 and CIFAR-10 results on full data are shown in Table \ref{tab:Results}, CIFAR-10 results on limited data in Figure \ref{fig:limited_CIFAR}.
For a fair comparison on STL-10, we adapted the official implementation of \cite{DISCO} which uses a stride of 1 in the initial convolutional layer by using stride 2 as in \cite{Cutout,E2STCNNs}. 

On CIFAR-10, we again demonstrate an increased sample efficiency of the invariant streams leading to superior performance for small dataset sizes (Figure \ref{fig:limited_CIFAR}). While the E(2)-invariant network is able to outperform the std. baseline in the full data regime, the scale-invariant network achieves subpar performance. We interpret the latter as a sign of less variance along the scale mode in this problem setup due to the rather small images contained in CIFAR-10. Thus, other invariances are more important to decide for the correct class. The scale-invariant stream seems to be too restrictive in the sense that it is unable to learn the full set of object invariants that the baseline architecture leverages to classify the objects. On the full data, II improves the performance for the scale-invariant architecture. However, the performance on the E(2)-invariant architecture is only on par. Nevertheless, \cite{Rath2} show that a rotation-invariant architecture using II outperforms the one with pooling in limited data regimes even when achieving slightly worse results on full data. We further investigate the advantages of II in Section \ref{sec:AblII}. 
Our combined network is able to achieve the best results for all data regimes by learning to combine the best information at each dataset size.
The triple stream outperforms the dual variant which indicates that the std. stream is able to capture important additional object invariances that are neglected by the restricted, invariant streams.

On STL-10, the E2-II network outperforms its counterpart without II significantly. Scale-II also slightly improves upon the baseline and II is clearly beneficial when combining multiple streams (see Table \ref{tab:AblMS}).
Our multi-stream network achieves a new state-of-the-art result (Table \ref{tab:Results}), even with constant number of parameters. This shows that incorporating prior knowledge about multiple transformations improves the performance of classification DNNs in the limited data domain, even for complex real-world datasets. Additionally, the results show that the features learned by each stream are complementary, preserved by our II layers and are effectively combined by our proposed multi-stream head. The multi-stream architecture successfully improves the sample complexity while raising the number of parameters -- hence increasing generalization.

\subsection{Ablation Studies}
\paragraph{Multi-Stream: Number of Parameters.}\label{sec:ConstantParams}
In addition to the full multi-stream architecture, we report the performance when keeping the number of parameters constant. Therefore, we shrink each stream by factor 2 or 3 for the dual or triple-stream network, respectively. The results are shown in Table \ref{tab:Results} marked with $^\star$. Limited data results for SVHN and CIFAR-10 are shown in Appendix \ref{sec:ConstantParams-sup}.

Our approach still achieves state-of-the-art performance on all datasets and in all data domains.
Combining multiple streams is beneficial, even with constant number of parameters. The dual-stream performs slightly better than the triple-stream. We believe this occurs since the individual streams of the triple-stream architecture are too thin, particularly in early layers.

Furthermore, we train a triple-stream architecture consisting of three standard streams on STL-10. We achieve a test error (TE) of 10.29\% which is worse than the single-stream networks with invariance. This shows that the enforced invariance plays a key role for our multi-stream networks. 

\paragraph{Multi-Stream: Importance of Invariant Integration.}\label{sec:AblII}
To quantify and demonstrate the importance of the II layer, we compare our multi-stream architecture including II to variants without II on STL-10. This includes a multi-stream architecture, where only pooling is used. The results in Table \ref{tab:AblMS} demonstrate that on a more complex classification task, in low-data regime (i.e. when the training data does not properly cover all variability present in the test data) the multi-stream approach works best when both streams use II rather than pooling.

Training a standard WRN augmented with random 90$^\circ$ rotations and scales $s \in [0.25, 1]$ on STL-10 achieves a TE of 21.80\%, which is clearly detrimental compared to the performance without those augmentations (12.08\%). Hence, layer-wise, guaranteed in- and equivariance play a key role in the improved sample complexity of our approach.

\section{\uppercase{Conclusion}}\label{sec:Conclusion}
In this contribution, we expanded II to scale transformations and showed its effectiveness on Scaled-MNIST. Since Scale-II using a WS is easier to optimize than the monomial variant, we applied it in a multi-stream DNN which besides scales includes a standard convolutional and a rotation-and-flip-invariant stream. This multi-stream DNN covers a variety of practically interesting use cases as shown by an improved sample complexity on SVHN and CIFAR-10 and new state-of-the-art results on STL-10 using only labeled data. We impose invariance to scales and rotation-and-flips in dedicated streams that also learn other global invariances and cover the remaining object invariances with a standard convolutional stream. This guarantees multiple invariances without suffering from the multiplicative increase when directly combining the groups.

Our framework is thought to leverage and honor prior knowledge. Therefore, it is focused on invariance guarantees, which may be rather restrictive in some cases. Specifically, invariance guarantees improve the sample complexity of DNNs leading to a performance boost, when training data is limited. In the large data domain, Vision Transformers \citep{ViT} with less geometrical constraints outperform conventional CNNs. Hence, our experiments focus on small-scale datasets. II is a general method that can be expanded beyond rotations and scales, but is restricted to transformations that can be modeled as (semi-) groups. Furthermore, we require an equivariant backbone before transferring to invariance. Nevertheless, the multi-stream network can in theory be enhanced with streams that achieve invariance without using II or G-Convs.

In the future, it is interesting to apply II to tasks where equivariance is helpful, e.g. to infer the pose in object detection. II could still be used for the parts of the network that benefit from invariance. On CIFAR-10 and SVHN, more sophisticated architectures and training methods than WRNs achieve a better performance \citep{SVHN1,SVHN2}. For a fair comparison, we stuck to WRNs in our experiments. Nevertheless, guaranteed invariances can generally be applied to many architectures in order to improve their sample complexity.

\subsubsection*{Acknowledgments}
We would like to thank the reviewers and our colleagues Lukas Enderich, Julia Lust and Paul Wimmer for their valuable remarks and contributions.
{
	\bibliography{egbib}
}

\appendix
\onecolumn

\section*{APPENDIX}
\section{\uppercase{Limited Data Studies with Constant Parameters}}\label{sec:ConstantParams-sup}
\begin{figure}[h]
	\centering
	\begin{minipage}[b]{.49\textwidth}
		\includegraphics[width=\textwidth]{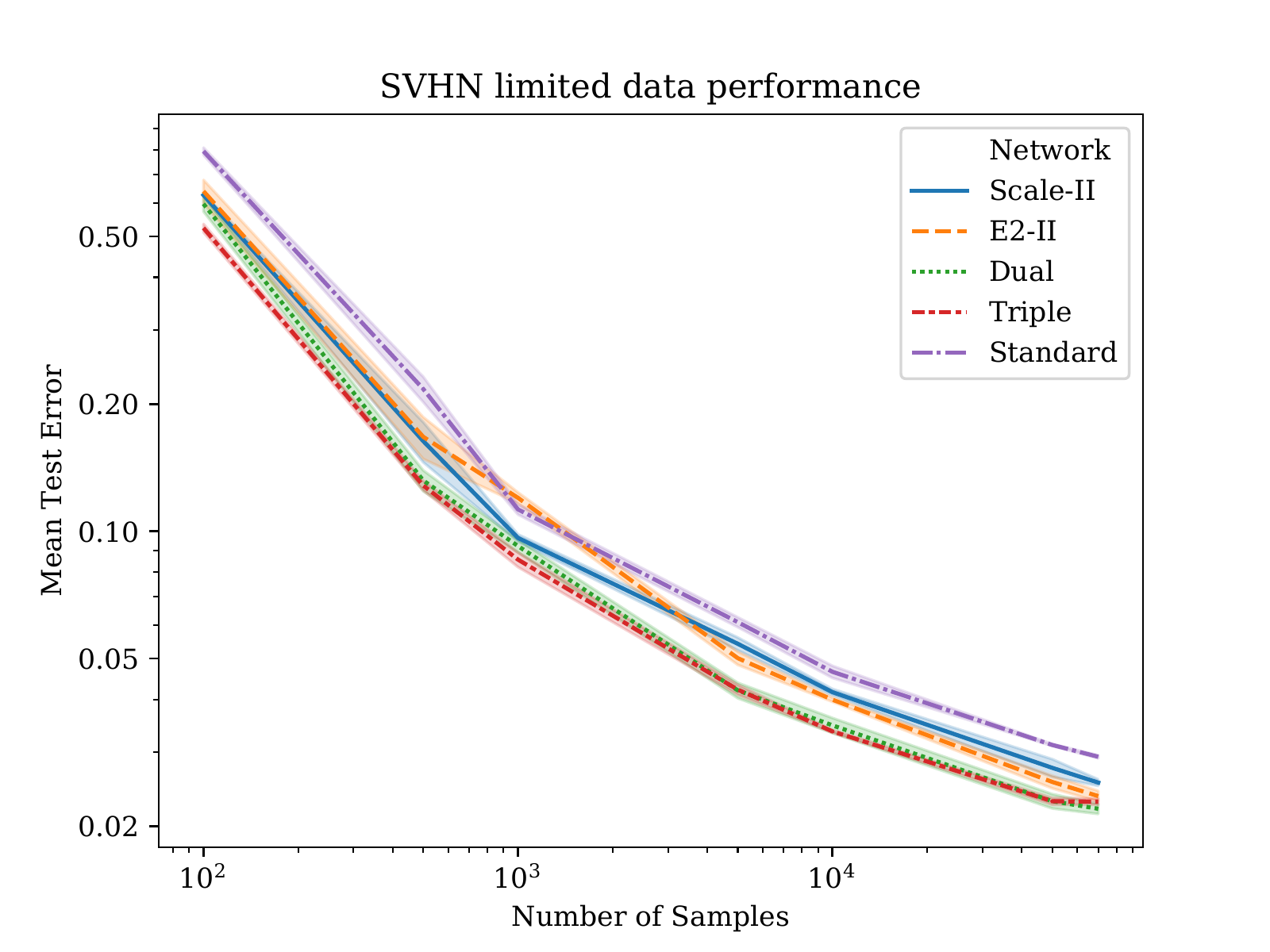}
		\caption{Log Test Error on limited subsets of SVHN with constant parameters.}
		\label{fig:limited_SVHNparams}
	\end{minipage}
	\hfill
	\begin{minipage}[b]{.49\textwidth}
		\centering
		\includegraphics[width=\textwidth]{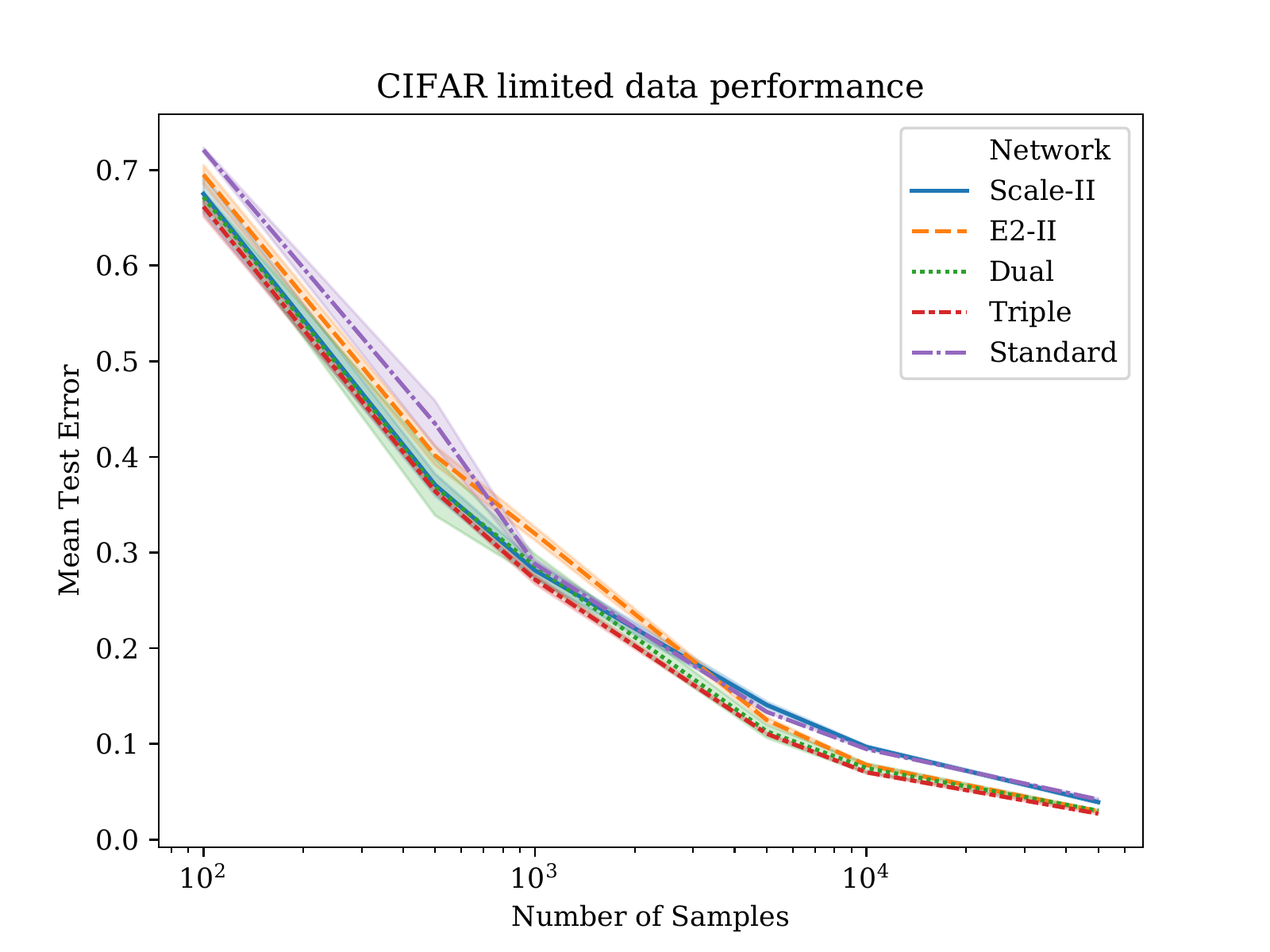}
		\captionof{figure}{Test Error on limited subsets of CIFAR-10 with constant parameters.}
		\label{fig:limited_CIFARparams}
	\end{minipage}
\end{figure}

In this section, we provide limited data graphics for SVHN (see Figure \ref{fig:limited_SVHNparams}) and CIFAR-10 (see Figure \ref{fig:limited_CIFARparams}) when using a constant number of parameters for our multi-stream network. The results on the full dataset are shown in the main paper in Table \ref{tab:Results}.
For the multi-stream networks with constant parameters, we observe a significant increase in the limited data domain and when using the full dataset compared to the baseline methods. While the dual stream performs slightly better on SVHN and STL-10, the triple stream achieves better performance in limited domains and on CIFAR. We conjecture that when the model capacity is limited the E(2)- and the scale-stream are able to learn most object invariances the standard stream would cover. Nevertheless, the standard stream adds the uncovered invariances leading to a slight performance boost. For the multi-stream networks with constant parameters, we used a naive down-scaling of the individual streams, i.e., the number of channels was simply divided by $\sqrt{2}$ or $\sqrt{3}$ per stream, respectively. We also did not further fine-tune hyper-parameters (HPs) for those down-scaled networks. With this straightforward approach, we already outperformed all single stream networks in limited data domains. In the future, further improvements could be achieved by investigating more sophisticated branching methods in order to enable a better allocation of how much capacity the network should spend for each invariant stream.

\section{\uppercase{Ablation Studies}}\label{sec:AblII-sup}
\subsection{Comparison of Multi-Stream Heads}\label{sec:Combination}
In this section, we compare different versions of our proposed Multi-Stream head to a concatenation head using the triple-stream architecture on STL-10. Map-\textit{Stream} describes which stream the respective outputs are mapped to, i.e., the respective stream uses an non-learnable identity mapping. When using Map-All, all streams use a learnable mapping at the same time. The results are shown in Table \ref{tab:MSHeads}. Mapping all features to the rotation stream and adding them via a weighted-sum leads to the best results and outperforms a naive concatenation of features along the channel dimension. 
\begin{table}[t]
	\centering
	\small
	\caption{Comparison of Multi-Stream heads on STL-10 using the triple stream architecture.}\label{tab:MSHeads}
	\begin{tabular}{cc} 
		Combination-Type & Test Error [\%] \\ 
		\midrule 
		Concat & 6.82 $\pm$ 0.11 \\
		Map-E(2) & \textbf{5.90} $\pm$ 0.05 \\
		Map-Scale & 7.21 $\pm$ 0.06 \\
		Map-Std & 6.65 $\pm$ 0.10 \\ 
		Map-All & 6.89 $\pm$ 0.06 \\
	\end{tabular} 
\end{table}

\subsection{Full End-to-End Training}\label{sec:AblFull}
As a further ablation, we conducted a "true" end-to-end training of our dual stream with random initialization and constant number of parameters on STL-10, where we achieve 7.86 \% Test Error (TE) without tuning any stream-specific hyper-parameters (HPs). We expect this to be slightly improvable when optimizing the stream-wise hyper-parameters.

Nevertheless, the true end-to-end training performs worse than our more intricate training procedure (6.38 \% TE). We conjecture, that optimizing each stream individually provides a good initialization point for the combination head and reduces the solution space the optimizer needs to consider. Overall, our learning strategy leads to a further improved sample complexity of the multi-stream architecture.

\section{\uppercase{Implementation Details \& Hyperparameters}}\label{sec:HPs}
We present details about the network architectures and the HPs needed to reproduce our results. We optimized the HPs using Bayesian optimization with hyperband (BOHB) \citep{BOHB}. For the baselines, we used the HPs reported in the respective papers or official implementations. If no validation split was pre-defined, we used a 80-20 train-validation split for the HP optimization. All networks were trained using a single (Scaled-MNIST, SVHN) or two (CIFAR-10, STL-10) NVIDIA GTX-1080 Ti GPUs. For SVHN, CIFAR-10 and STL-10 we used Wide-ResNets (WRNs) as baseline architectures \citep{WideResNet}.

For the E(2)-invariant network, we reduced the size of the final trivial representation to the same size used by the Standard- and Scale-WRN. For the original implementation, the final representation is multiplied by $\sqrt{|G|}$ where $|G|$ is the order of the group the last block is equivariant to \citep{E2STCNNs}. For the Scale-II WRNs, we added BatchNorm after the II layer to stabilize the gradients back-propagated through the division within the Scale-II. 

For the dual- and triple-stream architectures with constant parameters, we naively divided each channel size $C$ by $\lceil \frac{C}{\sqrt{2}} \rceil$ or $\lceil \frac{C}{\sqrt{3}} \rceil$, respectively. We did not further tune HPs for the smaller-sized streams.

We built upon the official code-bases of Sosnovik et al. (\cite{DISCO}, https://github.com/ISosnovik/disco), as well as Weiler and Cesa (\cite{E2STCNNs}, https://github.com/QUVA-Lab/e2cnn), which we both ported to Tensorflow v2.3. We used the models with a constant number of parameters and inserted the II layer to replace the spatial pooling operation. Specifically for DISCO, we had to thoroughly re-create all default settings from PyTorch in Tensorflow in order to recreate the performance. This included variable initializers, batch norm parameters and the Adam optimizer's HPs. Additionally, we had to use a custom-written \textit{grouped convolution} within the scale-equivariant convolution code rather than using \textit{tf.nn.conv2d} with a filter with reduced input-channels which implicitly also calculates a grouped convolution.

\subsection{Scaled-MNIST}
\begin{table}[t]
	\centering
	\caption{II-DISCO HPs on Scaled-MNIST. Parameters with $^\star$ were optimized using BOHB.}\label{tab:HPScaleMNIST}
	\small
	\begin{tabular}{ccc}
		HP & II-DISCO WS & II-DISCO Monomials \\ 
		\midrule 
		Batch Size & 128 & 128\\
		Learning Rate $^\star$ &  5e-3 & 1e-3\\  
		Weight Decay $^\star$ &  5e-7 & 5e-6 \\ 
		BatchNorm Decay $^\star$ & 0.01 & 5e-4 \\ 
		Dropout Rate $^\star$ & 0.1 & 0.7 \\
	\end{tabular} 
\end{table}

\begin{table}[t]
	\small
	\centering
	\caption{II-DISCO architecture on Scaled-MNIST.}
	\label{tab:IINetworkScaleMNIST}
	\begin{tabular}{cccccccc}
		Layer & Output Size & $C_{in}$ &  $C_{out}$ & $n_S$ & ReLU & BatchNorm & Dropout \\
		\midrule
		Upsampling & $ 56 \times 56$ & 1 & 1 & - & x & x & x \\
		Scale-LiftConv & $56 \times 56$ & 1 & 32 & 3 & \checkmark & \checkmark & x \\
		MaxPool &  $28 \times 28$ &  32 & 32 & 3 & x & x & x \\
		Scale-Conv1 & $28 \times 28$ & 32 & 63 & 3 & \checkmark & \checkmark & x \\
		MaxPool &  $14 \times 14$ &  63 & 63 & 3 & x & x & x \\
		Scale-Conv2 & $14 \times 14$ & 63 & 95 & 3 & \checkmark & \checkmark & x \\
		Scale-MaxProj & $14 \times 14$ & 95 & 95 & 1 & x & x & x \\
		Scale-II & $ 1 \times 1 $ & 95 & 95 & - & x & x & x \\
		Dense1 & - & 95 & 256  & - & \checkmark & \checkmark & \checkmark \\
		Dense2 & - & 256 & 10 & - & x & x & x \\
	\end{tabular}
\end{table}

On Scaled-MNIST, we used the CNN proposed by \cite{Sosnovik,DISCO} composed of three convolutional and two dense layers with an effective kernel size of $7$ and $n_S=4$ scales. We adapted it by using scale-II with $k=3$ instead of the spatial pooling layer that is a combination of average and max pooling. We trained our network for 60 epochs using the Adam optimizer \citep{Adam} with step-wise learning rate decay of $0.1$ after 20 and 40 epochs, $l_2$-regularization and data augmentation, where we artificially scaled the input with $s \in [0.5, 2]$. The network details can be found in Table \ref{tab:IINetworkScaleMNIST} and the used HPs in Table \ref{tab:HPScaleMNIST}.

\subsection{SVHN}
\begin{table}[t]
	\centering
	\caption{HPs used for SVHN experiments. Parameters with $^\star$ were optimized using BOHB.}\label{tab:HPSVHN}
	\small
	\begin{tabular}{ccc}
		HP & E(2)-II & Scale-II \\ 
		\midrule 
		Batch Size & 128 & 128 \\
		Learning Rate $^\star$ & 5e-3 & 0.2 \\
		Weight Decay $^\star$ & 5e-3 & 1e-4 \\
		Dropout Rate $^\star$ & 0.5 & 0.4 \\ 
	\end{tabular} 
\end{table}

\begin{table}[t]
	\centering
	\small
	\caption{Classification head HPs on SVHN.}\label{tab:HPCombSVHN}
	\begin{tabular}{ccc} 
		HP & Dual & Triple \\ 
		\midrule 
		Batch Size & 128 & 128 \\
		Learning Rate &  0.01 & 0.01 \\
		Weight Decay &  1e-4 & 1e-4  \\
	\end{tabular} 
\end{table}

\begin{table}[t]
	\centering
	\caption{E(2)-II-WRN16-4 architecture for SVHN. Each equivariant residual block consists of two consecutive convolution layers (e.g. E(2)-Conv1.1 and E(2)-Conv1.2) and a shortcut connection with a 1x1-convolution whenever the output size changes.} \label{tab:NetworkSVHN_E2}
	\small
	\begin{tabular}{ccccccccc}
		Layer & Output Size & $C_{in}$ &  $C_{out}$ & $n_r$ & $n_F$ & ReLU & BN & Dropout \\
		\midrule
		E(2)-LiftConv & $32 \times 32$ & 3 & 4 & 8 & 2 & \checkmark & \checkmark & x \\
		E(2)-Conv1.1 & $32 \times 32$ & 4 & 16 & 8 & 2 & \checkmark & \checkmark & x \\
		E(2)-Conv1.2 & $32 \times 32$ & 16 & 16 & 8 & 2 & \checkmark & \checkmark & \checkmark \\
		E(2)-Conv1.3 & $32 \times 32$ & 16 & 16 & 8 & 2 & \checkmark & \checkmark & x \\
		E(2)-Conv1.4 & $32 \times 32$ & 16 & 16 & 8 & 2 & \checkmark & \checkmark & \checkmark \\
		E(2)-Conv2.1 & $16 \times 16$ & 16 & 32 & 8 & 2 & \checkmark & \checkmark & x \\
		E(2)-Conv2.2 & $16 \times 16$ & 32 & 32 & 8 & 2 & \checkmark & \checkmark & \checkmark \\
		E(2)-Conv2.3 & $16 \times 16$ & 32 & 32 & 8 & 2 & \checkmark & \checkmark & x \\
		E(2)-Conv2.4 & $16 \times 16$ & 32 & 32 & 8 & 2 & \checkmark & \checkmark & \checkmark \\
		E(2)-Conv3.1 & $8 \times 8$ & 32 & 64 & 8 & 2 & \checkmark & \checkmark & x \\
		E(2)-Conv3.2 & $8 \times 8$ & 64 & 64 & 8 & 2 & \checkmark & \checkmark & \checkmark \\
		E(2)-Conv3.3 & $8 \times 8$ & 64 & 64 & 8 & 2 & \checkmark & \checkmark & x \\
		E(2)-Conv3.4 & $8 \times 8$ & 64 & 256 & 1 & 1 & \checkmark & \checkmark & \checkmark \\
		E(2)-II & - & 256 & 256 & - & - & x & x & x \\
		Dense & - & 256 & 10  & - & - & x & x & x \\
	\end{tabular}
\end{table}

\begin{table}[t]
	\centering
	\caption{Scale-II-WRN16-4 architecture for SVHN. Each equivariant residual block consists of two consecutive convolution layers (e.g. Scale-Conv1.1 and Scale-Conv1.2) and a shortcut connection with a 1x1-convolution whenever the output size changes.} \label{tab:NetworkSVHN_DISCO}
	\small
	\begin{tabular}{cccccccc}
		Layer & Output Size & $C_{in}$ &  $C_{out}$ & $n_S$ & ReLU & BN & Dropout \\
		\midrule
		Scale-LiftConv & $32 \times 32$ & 3 & 16 & 3 & \checkmark & \checkmark & x \\
		Scale-Conv1.1 & $32 \times 32$ & 16 & 64 & 3 & \checkmark & \checkmark & x \\
		Scale-Conv1.2 & $32 \times 32$ & 64 & 64 & 3 & \checkmark & \checkmark & \checkmark \\
		Scale-Conv1.3 & $32 \times 32$ & 64 & 64 & 3 & \checkmark & \checkmark & x \\
		Scale-Conv1.4 & $32 \times 32$ & 64 & 64 & 3 & \checkmark & \checkmark & \checkmark \\
		Scale-Conv2.1 & $16 \times 16$ & 64 & 128 & 3 & \checkmark & \checkmark & x \\
		Scale-Conv2.2 & $16 \times 16$ & 128 & 128 & 3 & \checkmark & \checkmark & \checkmark \\
		Scale-Conv2.3 & $16 \times 16$ & 128 & 128 & 3 & \checkmark & \checkmark & x \\
		Scale-Conv2.4 & $16 \times 16$ & 128 & 128 & 3 & \checkmark & \checkmark & \checkmark \\
		Scale-Conv3.1 & $8 \times 8$ & 128 & 256 & 3 & \checkmark & \checkmark & x \\
		Scale-Conv3.2 & $8 \times 8$ & 256 & 256 & 3 & \checkmark & \checkmark & \checkmark \\
		Scale-Conv3.3 & $8 \times 8$ & 256 & 256 & 3 & \checkmark & \checkmark & x \\
		Scale-Conv3.4 & $8 \times 8$ & 256 & 256 & 3 & \checkmark & \checkmark & \checkmark \\
		Scale-MaxProj & $8 \times 8$ & 256 & 256 & 1 &  x & x & x \\
		Scale-II & $1 \times 1$ & 256 & 256 & - &  x & \checkmark & x \\
		Dense & - & 256 & 10  & -  & x & x & x \\
	\end{tabular}
\end{table}
On the SVHN dataset, we used a WRN16-4 with pre-activation nonlinearites as baseline architecture. The E(2)-networks are equivariant to flips and $n_r=8$ rotations for each convolutional layer. The scale-convolutions use $n_S=3$ scales. We used E(2)-II with $k=3$, $n_r=8$ angles and $n_F=2$ flips using bi-linear interpolation where necessary. We applied Scale-II with $k=3$. We followed the training approach by \cite{WideResNet}. We used an SGD optimizer with Momentum $0.9$, trained for 160 epochs and reduced the learning rate by $0.1$ after 80 and 120 epochs. We did not apply any data augmentation.
The detailed architecture of the E(2)-II-WRN16-4 is shown in Table \ref{tab:NetworkSVHN_E2}, the architecture of the Scale-II-WRN16-4 in Table \ref{tab:NetworkSVHN_DISCO}. The used HPs are shown in Table \ref{tab:HPSVHN}.

For the classification head combining several invariant streams, we froze all trained convolutional and II layers and combined the individual streams via a learnable mapping and a weighted sum. We then trained those layer as well as the final dense layers. We used the same training settings as above, but divided the number of training epochs as well as all epoch-dependent HPs by 4. The HPs of the refinement training are listed in Table \ref{tab:HPCombSVHN}.

\subsection{CIFAR-10}
\begin{table}[t]
	\centering
	\caption{HPs on CIFAR-10. Parameters with $^\star$ were optimized using BOHB.}\label{tab:HPCIFAR}
	\small
	\begin{tabular}{ccc} 
		HP & E(2)-II & Scale-II \\ 
		\midrule 
		Batch Size & 96 & 128 \\
		Learning Rate $^\star$ & 5e-3 & 0.1 \\
		Weight Decay $^\star$ & 5e-3 & 5e-4 \\ 
		Dropout Rate $^\star$ & 0.1 & 0.2 \\ 
	\end{tabular} 
\end{table}

\begin{table}[t]
	\centering
	\caption{Classification head HPs on CIFAR-10.}\label{tab:HPCombCIFAR}
	\small
	\begin{tabular}{ccc}
		HP & Dual & Triple \\ 
		\midrule 
		Batch Size & 128 & 128 \\
		Learning Rate &  0.01 & 0.01  \\ 
		Weight Decay & 1e-3 & 1e-3 \\
	\end{tabular} 
\end{table}

\begin{table}[t]
	\centering
	\caption{E(2)-II-WRN28-10 architecture for CIFAR-10. Each equivariant residual block consists of two consecutive convolution layers (e.g. E(2)-Conv1.1 and E(2)-Conv1.2) and a shortcut connection with a 1x1-convolution whenever the output size changes.} \label{tab:NetworkCIFAR_E2}
	\small
	\begin{tabular}{ccccccccc}
		Layer & Output Size & $C_{in}$ &  $C_{out}$ & $n_r$ & $n_F$ & ReLU & BN & Dropout \\
		\midrule
		E(2)-LiftConv & $32 \times 32$ & 3 & 4 & 8 & 2 & \checkmark & \checkmark & x \\
		E(2)-Conv1.1 & $32 \times 32$ & 4 & 40 & 8 & 2 & \checkmark & \checkmark & x \\
		E(2)-Conv1.2 & $32 \times 32$ & 40 & 40 & 8 & 2 & \checkmark & \checkmark & \checkmark \\
		E(2)-Conv1.3 & $32 \times 32$ & 40 & 40 & 8 & 2 & \checkmark & \checkmark & x \\
		E(2)-Conv1.4 & $32 \times 32$ & 40 & 40 & 8 & 2 & \checkmark & \checkmark & \checkmark \\
		E(2)-Conv1.5 & $32 \times 32$ & 40 & 40 & 8 & 2 & \checkmark & \checkmark & x \\
		E(2)-Conv1.6 & $32 \times 32$ & 40 & 40 & 8 & 2 & \checkmark & \checkmark & \checkmark \\
		E(2)-Conv1.7 & $32 \times 32$ & 40 & 40 & 8 & 2 & \checkmark & \checkmark & x \\
		E(2)-Conv1.8 & $32 \times 32$ & 40 & 40 & 8 & 2 & \checkmark & \checkmark & \checkmark \\
		E(2)-Conv2.1 & $16 \times 16$ & 40 & 113 & 4 & 2 & \checkmark & \checkmark & x \\
		E(2)-Conv2.2 & $16 \times 16$ & 113 & 113 & 4 & 2 & \checkmark & \checkmark & \checkmark \\
		E(2)-Conv2.3 & $16 \times 16$ & 113 & 113 & 4 & 2 & \checkmark & \checkmark & x \\
		E(2)-Conv2.4 & $16 \times 16$ & 113 & 113 & 4 & 2 & \checkmark & \checkmark & \checkmark \\
		E(2)-Conv2.5 & $16 \times 16$ & 113 & 113 & 4 & 2 & \checkmark & \checkmark & x \\
		E(2)-Conv2.6 & $16 \times 16$ & 113 & 113 & 4 & 2 & \checkmark & \checkmark & \checkmark \\
		E(2)-Conv2.7 & $16 \times 16$ & 113 & 113 & 4 & 2 & \checkmark & \checkmark & x \\
		E(2)-Conv2.8 & $16 \times 16$ & 113 & 113 & 4 & 2 & \checkmark & \checkmark & \checkmark \\
		E(2)-Conv3.1 & $8 \times 8$ & 113 & 226 & 4 & 2 & \checkmark & \checkmark & x \\
		E(2)-Conv3.2 & $8 \times 8$ & 226 & 226 & 4 & 2 & \checkmark & \checkmark & \checkmark \\
		E(2)-Conv3.3 & $8 \times 8$ & 226 & 226 & 4 & 2 & \checkmark & \checkmark & x \\
		E(2)-Conv3.4 & $8 \times 8$ & 226 & 226 & 4 & 2 & \checkmark & \checkmark & \checkmark \\
		E(2)-Conv3.5 & $8 \times 8$ & 226 & 226 & 4 & 2 & \checkmark & \checkmark & x \\
		E(2)-Conv3.6 & $8 \times 8$ & 226 & 226 & 4 & 2 & \checkmark & \checkmark & \checkmark \\
		E(2)-Conv3.7 & $8 \times 8$ & 226 & 226 & 4 & 2 & \checkmark & \checkmark & x \\
		E(2)-Conv3.8 & $8 \times 8$ & 226 & 640 & 1 & 1 & \checkmark & \checkmark & \checkmark \\
		E(2)-II & - & 640 & 640 & - & - & x & x & x \\
		Dense & - & 640 & 10  & - & - & x & x & x \\
	\end{tabular}
\end{table}

\begin{table}[t]
	\centering
	\caption{Scale-II-WRN28-10 architecture for CIFAR-10. Each equivariant residual block consists of two consecutive convolution layers (e.g. Scale-Conv1.1 and Scale-Conv1.2) and a shortcut connection with a 1x1-convolution whenever the output size changes.} \label{tab:NetworkCIFAR_DISCO}
	\small
	\begin{tabular}{cccccccc}
		Layer & Output Size & $C_{in}$ &  $C_{out}$ & $n_S$ & ReLU & BN & Dropout \\
		\midrule
		Scale-LiftConv & $32 \times 32$ & 3 & 16 & 3 & \checkmark & \checkmark & x \\
		Scale-Conv1.1 & $32 \times 32$ & 16 & 160 & 3 & \checkmark & \checkmark & x \\
		Scale-Conv1.2 & $32 \times 32$ & 160 & 160 & 3 & \checkmark & \checkmark & \checkmark \\
		Scale-Conv1.3 & $32 \times 32$ & 160 & 160 & 3 & \checkmark & \checkmark & x \\
		Scale-Conv1.4 & $32 \times 32$ & 160 & 160 & 3 & \checkmark & \checkmark & \checkmark \\
		Scale-Conv1.5 & $32 \times 32$ & 160 & 160 & 3 & \checkmark & \checkmark & x \\
		Scale-Conv1.6 & $32 \times 32$ & 160 & 160 & 3 & \checkmark & \checkmark & \checkmark \\
		Scale-Conv1.7 & $32 \times 32$ & 160 & 160 & 3 & \checkmark & \checkmark & x \\
		Scale-Conv1.8 & $32 \times 32$ & 160 & 160 & 3 & \checkmark & \checkmark & \checkmark \\
		Scale-Conv2.1 & $16 \times 16$ & 160 & 320 & 3 & \checkmark & \checkmark & x \\
		Scale-Conv2.2 & $16 \times 16$ & 320 & 320 & 3 & \checkmark & \checkmark & \checkmark \\
		Scale-Conv2.3 & $16 \times 16$ & 320 & 320 & 3 & \checkmark & \checkmark & x \\
		Scale-Conv2.4 & $16 \times 16$ & 320 & 320 & 3 & \checkmark & \checkmark & \checkmark \\
		Scale-Conv2.5 & $16 \times 16$ & 320 & 320 & 3 & \checkmark & \checkmark & x \\
		Scale-Conv2.6 & $16 \times 16$ & 320 & 320 & 3 & \checkmark & \checkmark & \checkmark \\
		Scale-Conv2.7 & $16 \times 16$ & 320 & 320 & 3 & \checkmark & \checkmark & x \\
		Scale-Conv2.8 & $16 \times 16$ & 320 & 320 & 3 & \checkmark & \checkmark & \checkmark \\
		Scale-Conv3.1 & $8 \times 8$ & 320 & 640 & 3 & \checkmark & \checkmark & x \\
		Scale-Conv3.2 & $8 \times 8$ & 640 & 640 & 3 & \checkmark & \checkmark & \checkmark \\
		Scale-Conv3.3 & $8 \times 8$ & 640 & 640 & 3 & \checkmark & \checkmark & x \\
		Scale-Conv3.4 & $8 \times 8$ & 640 & 640 & 3 & \checkmark & \checkmark & \checkmark \\
		Scale-Conv3.5 & $8 \times 8$ & 640 & 640 & 3 & \checkmark & \checkmark & x \\
		Scale-Conv3.6 & $8 \times 8$ & 640 & 640 & 3 & \checkmark & \checkmark & \checkmark \\
		Scale-Conv3.7 & $8 \times 8$ & 640 & 640 & 3 & \checkmark & \checkmark & x \\
		Scale-Conv3.8 & $8 \times 8$ & 640 & 640 & 3 & \checkmark & \checkmark & \checkmark \\
		Scale-MaxProj & $8 \times 8$ & 640 & 640 & 1 &  x & x & x \\
		Scale-II & $1 \times 1$ & 640 & 640 & - &  x & \checkmark & x \\
		Dense & - & 640 & 10  & -  & x & x & x \\
	\end{tabular}
\end{table}
On CIFAR-10, we used a WRN28-10 as the baseline network. The E(2)-networks are equivariant to flips and $n_r=8$ rotations for the first residual block, $n_r=4$ rotations for the second and third residual blocks. The scale-convolutions use $n_S=3$ scales. We used E(2)-II with $k=3$, $n_r=4$ angles and $n_F=2$ flips. We applied Scale-II with $k=3$. We again followed the training approach by \cite{WideResNet}. We used an SGD optimizer with Momentum $0.9$, trained for 200 epochs and reduced the learning rate by $0.2$ after 60, 120 and 160 epochs. We used data augmentation with random pads-and-crops as well as random flips.
The detailed architecture of the E(2)-II-WRN28-10 is shown in Table \ref{tab:NetworkCIFAR_E2}, the architecture of the Scale-II-WRN28-10 in Table \ref{tab:NetworkCIFAR_DISCO}. The HPs can be found in Table \ref{tab:HPCIFAR}. We used the same approach for the combination heads as for SVHN where we divided all epochs by 4. The HPs of the re-training are shown in Table \ref{tab:HPCombCIFAR}.

\subsection{STL-10}
\begin{table}[t]
	\centering
	\caption{HPs on STL-10. Parameters with $^\star$ were optimized using BOHB.}\label{tab:HPSTL}
	\small
	\begin{tabular}{ccc} 
		HP & E(2)-II & Scale-II \\ 
		\midrule 
		Batch Size & 96 & 96 \\
		Learning Rate $^\star$ & 2e-3 & 0.02 \\
		Weight Decay $^\star$ & 1e-2 & 1e-3 \\ 
		Dropout Rate $^\star$ & 0.1 & 0.25 \\ 
	\end{tabular} 
\end{table}

\begin{table}[t]
	\centering
	\caption{Classification head HPs on STL-10.}\label{tab:HPCombSTL}
	\small
	\begin{tabular}{ccc}
		HP & Dual & Triple \\ 
		\midrule 
		Batch Size & 128 & 128 \\
		Learning Rate &  0.01 & 0.01  \\ 
		Weight Decay & 1e-4 & 1e-4 \\
	\end{tabular} 
\end{table}

\begin{table}[t]
	\centering
	\caption{E(2)-II-WRN16-8 architecture for STL-10. Each equivariant residual block consists of two consecutive convolution layers (e.g. E(2)-Conv1.1 and E(2)-Conv1.2) and a shortcut connection with a 1x1-convolution whenever the output size changes.} \label{tab:NetworkSTL_E2}
	\small
	\begin{tabular}{ccccccccc}
		Layer & Output Size & $C_{in}$ &  $C_{out}$ & $n_r$ & $n_F$ & ReLU & BN & Dropout \\
		\midrule
		E(2)-LiftConv & $96 \times 96$ & 3 & 4 & 8 & 2 & \checkmark & \checkmark & x \\
		E(2)-Conv1.1 & $48 \times 48$ & 4 & 32 & 8 & 2 & \checkmark & \checkmark & x \\
		E(2)-Conv1.2 & $48 \times 48$ & 32 & 32 & 8 & 2 & \checkmark & \checkmark & \checkmark \\
		E(2)-Conv1.3 & $48 \times 48$ & 32 & 32 & 8 & 2 & \checkmark & \checkmark & x \\
		E(2)-Conv1.4 & $48 \times 48$ & 32 & 32 & 8 & 2 & \checkmark & \checkmark & \checkmark \\
		E(2)-Conv2.1 & $24 \times 24$ & 32 & 90 & 4 & 2 & \checkmark & \checkmark & x \\
		E(2)-Conv2.2 & $24 \times 24$ & 90 & 90 & 4 & 2 & \checkmark & \checkmark & \checkmark \\
		E(2)-Conv2.3 & $24 \times 24$ & 90 & 90 & 4 & 2 & \checkmark & \checkmark & x \\
		E(2)-Conv2.4 & $24 \times 24$ & 90 & 90 & 4 & 2 & \checkmark & \checkmark & \checkmark \\
		E(2)-Conv3.1 & $12 \times 12$ & 90 & 362 & 1 & 2 & \checkmark & \checkmark & x \\
		E(2)-Conv3.2 & $12 \times 12$ & 362 & 362 & 1 & 2 & \checkmark & \checkmark & \checkmark \\
		E(2)-Conv3.3 & $12 \times 12$ & 362 & 362 & 1 & 2 & \checkmark & \checkmark & x \\
		E(2)-Conv3.4 & $12 \times 12$ & 362 & 512 & 1 & 1 & \checkmark & \checkmark & \checkmark \\
		E(2)-II & - & 512 & 512 & - & - & x & x & x \\
		Dense & - & 512 & 10  & - & - & x & x & x \\
	\end{tabular}
\end{table}

\begin{table}[t]
	\centering
	\caption{Scale-II-WRN16-8 architecture for STL-10. Each equivariant residual block consists of two consecutive convolution layers (e.g. Scale-Conv1.1 and Scale-Conv1.2) and a shortcut connection with a 1x1-convolution whenever the output size changes.} \label{tab:NetworkSTL_DISCO}
	\small
	\begin{tabular}{cccccccc}
		Layer & Output Size & $C_{in}$ &  $C_{out}$ & $n_S$ & ReLU & BN & Dropout \\
		\midrule
		Scale-LiftConv & $96 \times 96$ & 3 & 32 & 3 & \checkmark & \checkmark & x \\
		Scale-Conv1.1 & $48 \times 48$ & 32 & 128 & 3 & \checkmark & \checkmark & x \\
		Scale-Conv1.2 & $48 \times 48$ & 128 & 128 & 3 & \checkmark & \checkmark & \checkmark \\
		Scale-Conv1.3 & $48 \times 48$ & 128 & 128 & 3 & \checkmark & \checkmark & x \\
		Scale-Conv1.4 & $48 \times 48$ & 128 & 128 & 3 & \checkmark & \checkmark & \checkmark \\
		Scale-Conv2.1 & $24 \times 24$ & 128 & 256 & 3 & \checkmark & \checkmark & x \\
		Scale-Conv2.2 & $24 \times 24$ & 256 & 256 & 3 & \checkmark & \checkmark & \checkmark \\
		Scale-Conv2.3 & $24 \times 24$ & 256 & 256 & 3 & \checkmark & \checkmark & x \\
		Scale-Conv2.4 & $24 \times 24$ & 256 & 256 & 3 & \checkmark & \checkmark & \checkmark \\
		Scale-Conv3.1 & $12 \times 12$ & 256 & 512 & 3 & \checkmark & \checkmark & x \\
		Scale-Conv3.2 & $12 \times 12$ & 512 & 512 & 3 & \checkmark & \checkmark & \checkmark \\
		Scale-Conv3.3 & $12 \times 12$ & 512 & 512 & 3 & \checkmark & \checkmark & x \\
		Scale-Conv3.4 & $12 \times 12$ & 512 & 512 & 3 & \checkmark & \checkmark & \checkmark \\
		Scale-MaxProj & $12 \times 12$ & 512 & 512 & 1 &  x & x & x \\
		Scale-II & $1 \times 1$ & 512 & 512 & - &  x & \checkmark & x \\
		Dense & - & 512 & 10  & -  & x & x & x \\
	\end{tabular}
\end{table}

On STL-10, we used a WRN16-8 as the baseline network. The E(2)-networks are equivariant to flips as well as $n_r=8$ rotations for the first residual block, $n_r=4$ rotations for the second and $n_r=1$ rotations for the third residual block. The scale-convolutions use $n_S=3$ scales. We used E(2)-II with $k=3$, $n_r=1$ angles and $n_F=2$ flips and Scale-II with $k=3$. We used an SGD optimizer with Nesterov Momentum $0.9$, trained for 1000 epochs and reduced the learning rate by $0.2$ after 300, 400, 600 and 800 epochs. We used data augmentation with random pads-and-crops, flips and Cutout \citep{Cutout}.
The detailed architecture of the E(2)-II-WRN16-8 is shown in Table \ref{tab:NetworkSTL_E2}, the architecture of the Scale-II-WRN16-8 in Table \ref{tab:NetworkSTL_DISCO}. We adapted the official implementation of the Scale-networks to use a stride of 2 instead of 1 in the initial residual block -- as done by \cite{E2STCNNs,Cutout}. The used HPs can be found in Table \ref{tab:HPSTL}. We used the same approach for the combination heads as for SVHN and CIFAR-10, but this time divided all epochs by 10. The HPs of the re-training can be found in Table \ref{tab:HPCombSTL}.

\section{\uppercase{Broader Societal Impact}}\label{sec:Impact}
Our proposed method increases the sample-efficiency of DNNs for classification tasks. It can thus be used to train supervised models when training data is scarce or expensive to label, i.e., in fields such as medical imaging or autonomous driving. 
Since the methods proposed in this paper are quite general and not bound to a specific application, they can be used for any type of classification network processing images. This includes potentially harmful applications. Even without harmful intents, training on biased datasets as well as misclassifications can lead to unintended negative consequences such as a wrong medical treatment. 

The authors firmly renounce using our proposed methods with any harmful intents. When applying our methods, it is inevitable to monitor the decisions made by the network specifically based on ethical standards, such that the classifier does not decide in a harmfully biased way. Impactful decisions based on the output of DNNs, e.g. deciding on a medication or driving autonomously, need to be carefully supervised by human experts and/or redundant systems where appropriate.

\section{\uppercase{Licenses}}\label{sec:License}
We implemented all networks using Tensorflow v2.3 which is licensed under Apache 2.0 \citep{tensorflow2015-whitepaper}. We built upon the code-bases of \cite{DISCO} which is under MIT license and \cite{E2STCNNs} distributed under BSD Clear license. We ported both frameworks from PyTorch to Tensorflow v2.3. The used datasets Scaled-MNIST \citep{MNIST-Scale}, SVHN \citep{SVHN}, CIFAR-10 \citep{CIFAR-10} and STL-10 \citep{STL10} are not under license.

\section{\uppercase{Amount of Compute}}\label{sec:Compute}
For our experiments on Scaled-MNIST and SVHN, we used a single Nvidia 1080 GTX Ti. On CIFAR-10 and STL-10, we used two NVIDIA 1080 GTX Tis. 
For Scaled-MNIST, we performed $6$ runs \`a $8$ sizes, $2$ architectures, which equates to $96$ runs. In total, the GPU train time took $\approx 50$ hours.
For SVHN, we evaluated $3\cdot 2 + 7 \cdot 3 \cdot 7 = 153$ runs with $\approx 1800$ hours train time, for CIFAR-10 $3 \cdot 2 + 6 \cdot 3 \cdot 7 = 132$ runs with $\approx 3600$ hours train time and $1 \cdot 3 \cdot 9 = 27$ runs with $\approx 650$ hours train time for STL-10. We took the average train time for all architectures on each dataset to calculate those numbers. In total, we estimate the GPU time for the main results to $50 + 1800 + 2\cdot 3600 + 2\cdot 650 = 10350$ hours. 

\end{document}